\documentclass[11pt]{article} 
\usepackage{rldmsubmit,palatino}
\usepackage{graphicx}
\usepackage{amsmath}
\usepackage{subcaption}
\usepackage{placeins}
\usepackage{svg}
\usepackage{multicol}
\usepackage{hyperref}

\title{Utilizing Reinforcement Learning for Bottom-Up part-wise Reconstruction of 2D Wire-Frame Projections}

\author{
Julian Ziegler\\%
Laboratory for Biosignal Processing \\
Leipzig University of Applied Sciences\\
Leipzig, Germany \\
\texttt{julian.ziegler@htwk-leipzig.de} \\
\AND
Patrick Frenzel \\
Laboratory for Biosignal Processing \\
Leipzig University of Applied Sciences\\
Leipzig, Germany \\
\texttt{patrick.frenzel@htwk-leipzig.de} \\
\And
Mirco Fuchs \\
Laboratory for Biosignal Processing \\
Leipzig University of Applied Sciences\\
Leipzig, Germany \\
\texttt{mirco.fuchs@htwk-leipzig.de} \\
}

\begin{document}

\maketitle

\begin{abstract}

This work concerns itself with the task of reconstructing all edges of an arbitrary 3D wire-frame model projected to an image plane. We explore a bottom-up part-wise procedure undertaken by an RL agent to iteratively segment and reconstruct these 2D multipart objects. 
The environment's state is represented as a four-colour image, where different colours correspond to background, a target edge, a reconstruction line, and the overlap of both.
At each step, the agent can transform the reconstruction line within a four-dimensional action space or terminate the episode using a specific termination action.

To investigate the impact of reward function formulations, we tested episodic and incremental rewards, as well as combined approaches.
Empirical results demonstrated that the latter yielded the most effective training performance.
To further enhance efficiency and stability, we introduce curriculum learning strategies.
First, an action-based curriculum was implemented, where the agent was initially restricted to a reduced action space, being able to only perform three of the five possible actions, before progressing to the full action space.
Second, we test a task-based curriculum, where the agent first solves a simplified version of the problem before being presented with the full, more complex task.

This second approach in particular produced promising results, as the agent not only successfully transitioned from learning the simplified task to mastering the full task, but in doing so gained significant performance.
In conclusion, this study demonstrates the potential of an iterative RL wire-frame reconstruction in two dimensions.
By combining optimized reward function formulations with curriculum learning strategies, we achieved significant improvements in training success.
The proposed methodology provides an effective framework for solving similar tasks and represents a promising direction for future research in the field.

\end{abstract}

\keywords{PPO, Wire-Frame, Image-based RL, Learning Curriculum}

\startmain

\section{Introduction}
Reinforcement Learning (RL) has gained attention for its potential to provide intuitive and efficient solutions to complex problems. Unlike supervised or unsupervised learning, RL enables agents to learn directly from interactions with the environment without explicit supervision.
In this work, we investigate the use of an RL agent to perform part-wise reconstruction of two-dimensional wire-frames within a projection.
Such an approach may be applicable to image segmentation problems where all visible parts of 3D wire-frame objects as, for example, scaffoldings need to be identified. Assuming all composing parts of the object are known, this imitates a rather intuitive solving strategy, which a human expert would be expected to pursue.
The agent is able to iteratively identify target edges though transformation action within the environment (i.e. the projection of a wire-frame to an image corresponding to an image recorded with a camera), with the goal of fitting a reconstruction line to one of the targets.
\section{Related Work}
We provide a brief overview of current and past research related to our presented work.
A wire-frame model typically represents an arbitrary object by adjoining vertices and edges.
Reconstructing these wire-frames from an input image greatly increases usable information for downstream processes.
Huang et al. \cite{huang_learning_2018} showed reconstructing a 3D-model from an image using deep learning, utilizing detected edges and vertices to build wire-frames.
Even closer to our own work, Xue et al. \cite{xue_holistically-attracted_2023} used deep learning techniques to reconstruct 2D wire-frames of man-made objects contained in an image. They used a top-down approach, first detecting joints and lines before combining them in a second stage.
A similar process was recently used by Pautrat et al. \cite{pautrat_deeplsd_2023}.
The method proposed in this work differs fundamentally from these approaches, as we follow a bottom-up procedure based on RL to identify wire-frames.

Reinforcement Learning has been successfully used for image processing tasks like segmentation or object detection \cite{casanova_reinforced_2020}.
Srikishan et al. \cite{srikishan_reinforcement_2024} used RL to decide whether detected objects should be passed to more powerful detectors to receive more precise results.
Similar to our own work, Wang et al. \cite{wang_outline_2018} used RL to segment object boundaries. In contrast to our work, however, their work is based on a general image dataset. 
Jie et al. \cite{jie_tree-structured_2017} present an agent based object detection in scenes where multiple objects are present. They utilized a tree structure that allows the agent to efficiently discover all objects in a given frame with minimal actions used.
Contrary to all mentioned works, we use the agent itself to transform the visual state signal, therefore it is able to learn actions based on easy to implement visual triggers.

We will propose to train the agent on a curriculum built from different settings of the environment. 
From a human perspective, training on easier, less complex tasks seems intuitive. 
The theoretical groundwork for curriculum learning in a machine learning context was laid out by Bengio et al. \cite{bengio-curriculum}. They hypothesize it to be a special form of the continuation method, a strategy often used for global optimization of non-convex functions.
More recently, Florensa et al. \cite{florensa_reverse_2018} show an RL agent can make use of curricula during the learning phase.
However, they define subgoals to be specific states within the environment, which differs from our own approach.
\section{Problem Setting}

\begin{figure}[h!]
    \centering
    \includegraphics[width=\textwidth,trim={0cm 0cm 0cm 0cm}, clip]{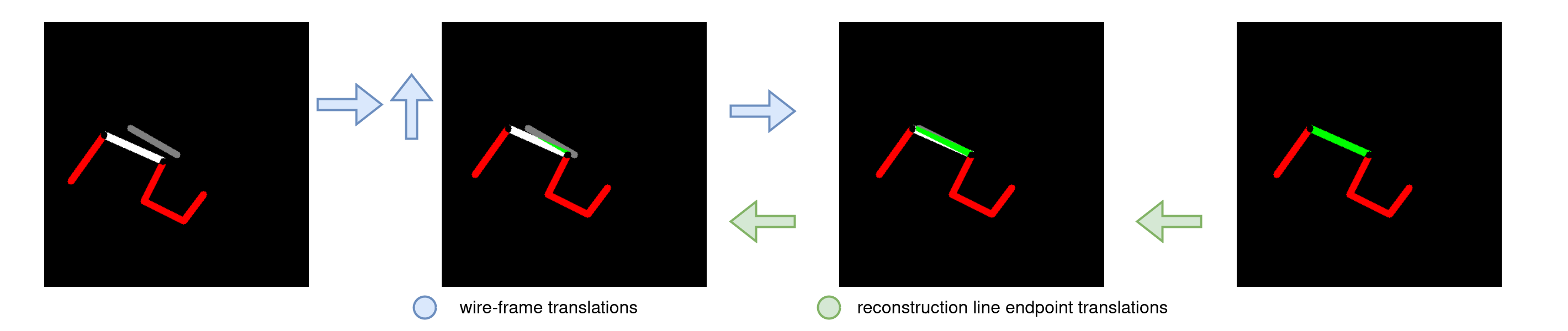}
    
\caption{Example Episode with wire-frame translations (blue) and reconstruction line transformations (green) highlighted. For more in-depth visualizations, please visit our \protect\href{https://github.com/juliantziegler/RL_Wire-Frame_Reconstruction}{GitHub Repository}.}

    \label{fig:fatm_plots}
\end{figure}

We implement the task of reconstructing arbitrarily generated 2D wire-frames in an episodic RL environment.
Each episode consists of finding one of the target edges constituting the entire 2D frame.
The wire-frame is encoded as white pixels on a 30 by 30 pixel black background image. 
To solve the environment, the agent must perfectly align one of the target edges and the current reconstruction line, which is colour-coded gray.
The target image is placed into the larger 60 by 60 state space image, where one of the reconstruction lines endpoints is anchored to the midpoint.
This was done to reduce the state representation space, as the reconstruction line is always centred, and therefore can be in far fewer states.
An overlap of target and reconstruction line is represented by green pixels. If some target edges have already been detected in previous episodes, their corresponding pixels are coloured red. The task of an agent during one episode can be interpreted as minimizing the number of gray and white pixels while maximizing the number of green pixels, constrained by the wire-frame target and based on linear transformations of the reconstruction line. An example of an episode of a reconstruction procedure is shown in Fig. \ref{fig:fatm_plots}.

The agent can transform the state with an up to four dimensional action space to better align the target and reconstruction line.
Two of these translate the wire-frame, one pixel at a time, along both axes, while the other two can shift the non-anchored endpoint of the reconstruction line in the 2D space, thereby allowing scaling and rotation of the line. Overall, performing these actions allows the agent to align the reconstruction line with the wire-frame in an iterative procedure. 
Additionally, a fixate action is implemented to terminate an episode. It allows the agent to stop the current reconstruction prematurely if no further action is required to align it with the target. Otherwise, episodes are ended after 200 steps.

\section{Methodology}
\paragraph{Environment and Action Space Configurations:}
We previously outlined the fundamental aspects of the environment and its represented goal, e.g. the image state signal or the theoretical action space. Here, we detail the methods and variations used in our studies.

First, we implement two variations of the action space denoted as SAT and FAT. 
The Simple Action Space (SAT) only implements the translation actions for the target wire-frame and the fixate action, thereby not allowing the transformation of the reconstruction line.
For perfect alignment to be possible for this limited action space, at reset the reconstruction line is generated to resemble a randomly chosen target edge. In other words, a perfect reconstruction can be found without any rotation but solely based on wire-frame translations.
The Full Action Space (FAT) on the other hand implements all mentioned action dimensions, thus effectively allowing for all linear transformations (translate, scale and rotate) of the reconstruction line.\\
Second, we utilize two variants of the reconstruction task, i.e. single detection mode and multi detection mode. As outlined earlier, the intended task of the agent is to iteratively find all target edges in the generated scene, thereby reconstructing the wire-frame. To simplify this original problem, single detection mode resets the entire scene after each episode. As a consequence, the reconstruction procedures for all episodes are independent of each other since previously identified edges are not carried over to subsequent episodes, thereby effectively reducing task complexity. In contrast, the multi detection mode refers to the full task, i.e. reconstructing the whole wire-frame through consecutive line detections.

All described environments were implemented using OpenAi`s Gym Framework \cite{gym}.
It is used throughout the experiments outlined below, examining both reward function design and learning curricula.
\paragraph{Reward Functions:}
In episodic reinforcement learning, rewards can be given after each step or after the completion of an episode. 
We propose a step-wise reward $r_t$ and an episodic return $E_t$ to be given when terminating the episode, and analyse performance against each other.
Furthermore, we combine both schemes and apply clipping procedures.

From step $s_{t-1}$ to the next step $s_{t}$, the reward $r_t$ is formed by comparing the IoU in both states.
This can be done efficiently, as the colours correspond to True Positive (green), false positive (gray), false negative (white) and true negative (black).
The step-wise reward itself is defined as
\begin{equation}
    r_t = \begin{cases}
    \begin{aligned}
    0.1 & \quad IoU_t > IoU_{t-1} \\
    \text{-}0.2 & \quad IoU_t < IoU_{t-1} \\
    \text{-}0.01 & \quad IoU_t = IoU_{t-1} = 0 \\
    0 & \quad IoU_t = IoU_{t-1} \neq 0
    \end{aligned}
    \end{cases}
    \label{eq:methods-env-rew-stepreward}
\end{equation}
The negative reward earned by reducing the IoU is of higher absolute value than the positive reward for gaining IoU.
This was found to stop oscillation between two states, which for the same absolute value results in a net episodic return of 0. The episodic return resulting from reward $r_t$ is given by $R_t = \sum_{i=0}^{t}{\gamma^{t-i} r_i}$, where $\gamma$ denotes the discount factor.\\ 
At the end of an episode, we propose to form the reward $E_t$ by evaluating the similarity of the reconstruction line to all target edges present in the environment. For simplicity, this is done based on Euclidean distances between endpoints of reconstruction and target lines.
The episodic reward is therefore defined as
\begin{equation}
    E_t = \min \left[-\mu; \mu \left(1 - \frac{2}{D_t} \min_i \left(||p^1  - q_{i}^c||_2 + ||p^2 - q_i^{\overline{c}}||_2 \right)\right)\right] ,
    \label{eq: methods-env-rew-distancetrigger}
\end{equation}
with $||\cdot||_2$ the $\ell_2$-norm, $p^1$ and $p^2$ are the endpoints of the reconstruction line, $q_i^c$ is the endpoint of target edge $i$ closest to endpoint $p^1$ and $q_i^{\overline{c}}$ the other endpoint of target edge $i$.  $D_t$ and $\mu$ are hyperparameters, the former determining the slope of the function and the latter determining its range. Total episodic return $G_t$ defines as $G_t = R_t + E_t$. In the experiments outlined in Sec. Experiments below, we define $R_t=0$ for the sparse reward scheme and $E_t=0$ for the incremental reward scheme.\\
Reflecting on the task at hand, it is obvious that the agent, while indeed aligning perfectly, will not collect the same return in each instantiation.
This is due to the step-wise reward $r_t$ implicitly favouring more IoU -increasing steps taken, as this increases return regardless of alignment.
Though somewhat mitigated by the $\gamma$ -hyperparameter of the PPO Algorithm, it could still pose an obstacle in learning generally applicable, optimal actions.
Designed to counteract this, we test two clipping methods. 
First, we simply clip the return of an episode at $+ \mu$, i.e. return is given by \begin{equation}G_t^{\textnormal{clip}} = \min\left(G_t, \mu\right).\end{equation} As a more sophisticated approach, we clip the return to the episodic reward $E_t$, if this is indeed positive, following the equation
\begin{equation}
    G_t^\textnormal{clip+} =\begin{cases}
    \begin{aligned}
     \min\left(G_t, E_t\right) & \quad E_t > 0 \\
    G_t & \quad \textnormal{otherwise} \\
    \end{aligned}
    \end{cases}
\end{equation}
This allows the agent to learn to overlap the edges initially, but also lays a bigger focus on the actual fit of the edges to determine the return.
\paragraph{Learning Curricula:}
\label{sec:curricula}
To examine the applicability of curriculum learning to the implemented environment and goal, we propose two learning curricula. 
First, we utilize both the SAT and FAT variants of the environment to build an action-based curriculum.
The agent at first is only required to learn three actions, which will later also be applicable to the FAT environment. 
We use the single detection mode throughout this curriculum to remove unnecessary variability. 
Secondly, we use single and multi detection mode to construct a difficulty curriculum.
The agent initially learns in the single detection environment, before transferring to the multi detection variant.
FAT action space is used throughout the experiment.
In both cases, a 30/70 split is used, meaning the agent trains in the first environment for 30\% of the overall time steps, before moving to the other for 70\% of training steps.
For comparisons with ordinary direct training, agents are trained for the full amount of time steps on the second environment.

\paragraph{Experiments:}
\label{sec:experiment}

At first, we will conduct an experiment to asses the proposed reward functions. We use (a) a sparse reward setting ($G_t=E_t$) where reward only depends on the reconstruction result at the end of an episode, (b) an only-incremental reward setting ($G_t=R_t$) where reward is only driven by IoU increments throughout the episode, (c) a combination of a) and b), (d) the reward scheme similar to c) but clipped to $+\mu$ ($G_t=G_t^\textnormal{clip}$) and, likewise, (e) with return clipped to positive $E_t$ ($G_t=G_t^\textnormal{clip+}$). We always use the FAT version of the environment in single detection mode and limit the size of the wire-frame to three edges. The $\mu$ parameter in $E_t$ was set to $5$. The agent is trained for each reward formulation for 10 million time steps in the environment.
Results are based on an aggregation of 10 successful runs.

The second experiment aims to investigate the proposed action-based and difficulty-based curricula. The combined reward function, where return in given by $G_t = R_t + E_t$, is used here, as it was deemed the most performant in the previous experiment.
Using the action curriculum, we train for 10 million timesteps. For the difficulty curriculum, we increase this to 20 million for more conclusive results.
The presented results will be the mean of 10 conducted runs.
\paragraph{Experimental Setup:}
We use the PPO algorithm first introduced by Schulman et al. \cite{schulman_proximal_2017}, with its originally proposed clipped surrogate objective
Our implementation is based upon the StableBaselines3 GitHub repository \cite{stable-baselines3}.
To extract features from the image state, we employ the neural network first used Mnih et al. \cite{mnih_human-level_2015} and often known as NatureCNN. 
Value and Action Functions are separately modelled using a head with two 64-unit layers each and ReLU activations.
A learning rate of $2.5 \cdot 10^{-4}$ with linear decay was utilized.
All experiments were run on 4 logical cores of an Intel Core i7 7700k CPU and an Nvidia GeForce GTX 1080 Ti GPGPU.
\section{Results}
\begin{figure}[h!]
    \begin{subfigure}[b]{0.27\textwidth}
        \includegraphics[width=\textwidth]{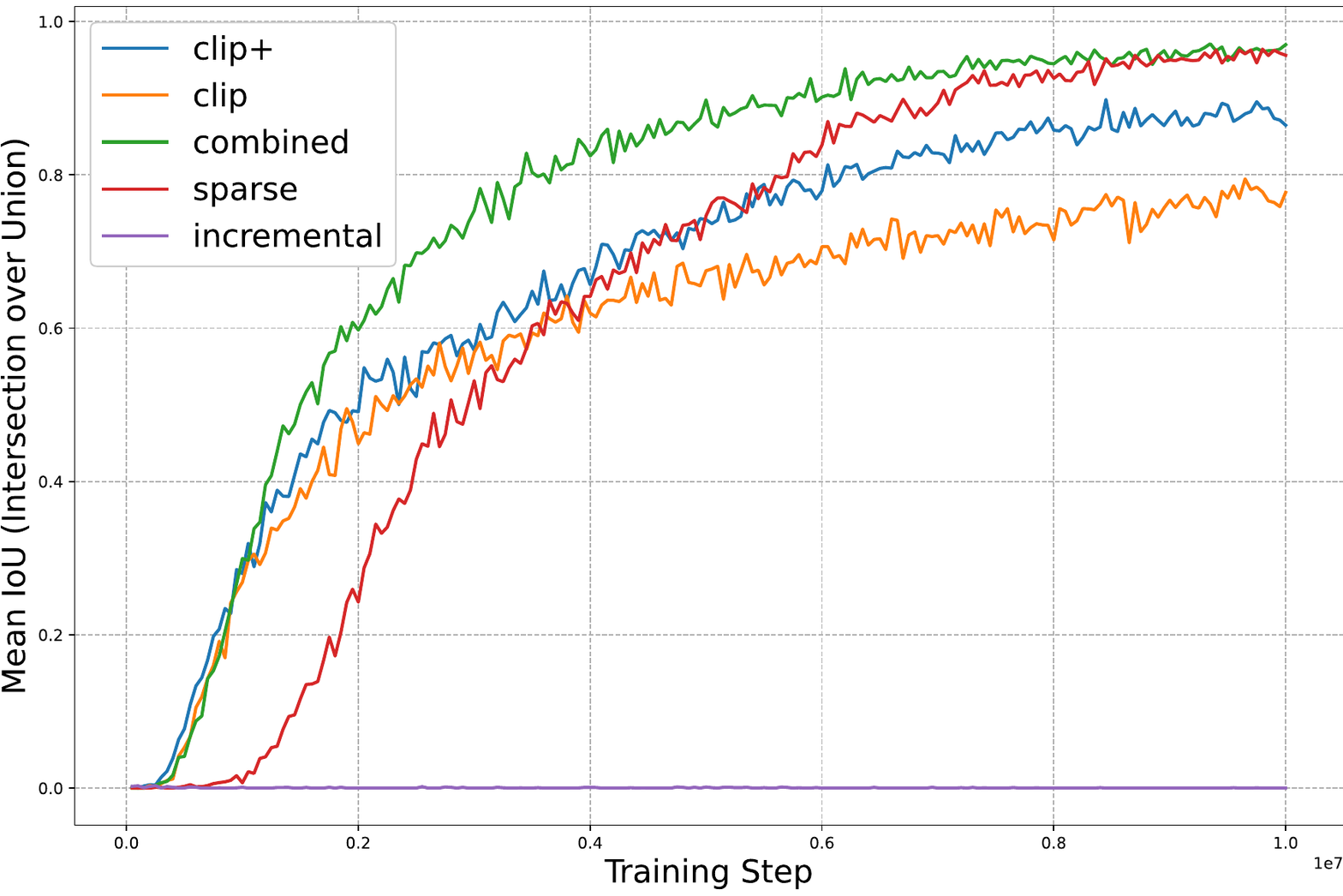}
        \caption{Reward function results}
        \label{fig:rewards}
    \end{subfigure}
    \hfill
    \begin{subfigure}[b]{0.27\textwidth}
        \includegraphics[width=\textwidth]{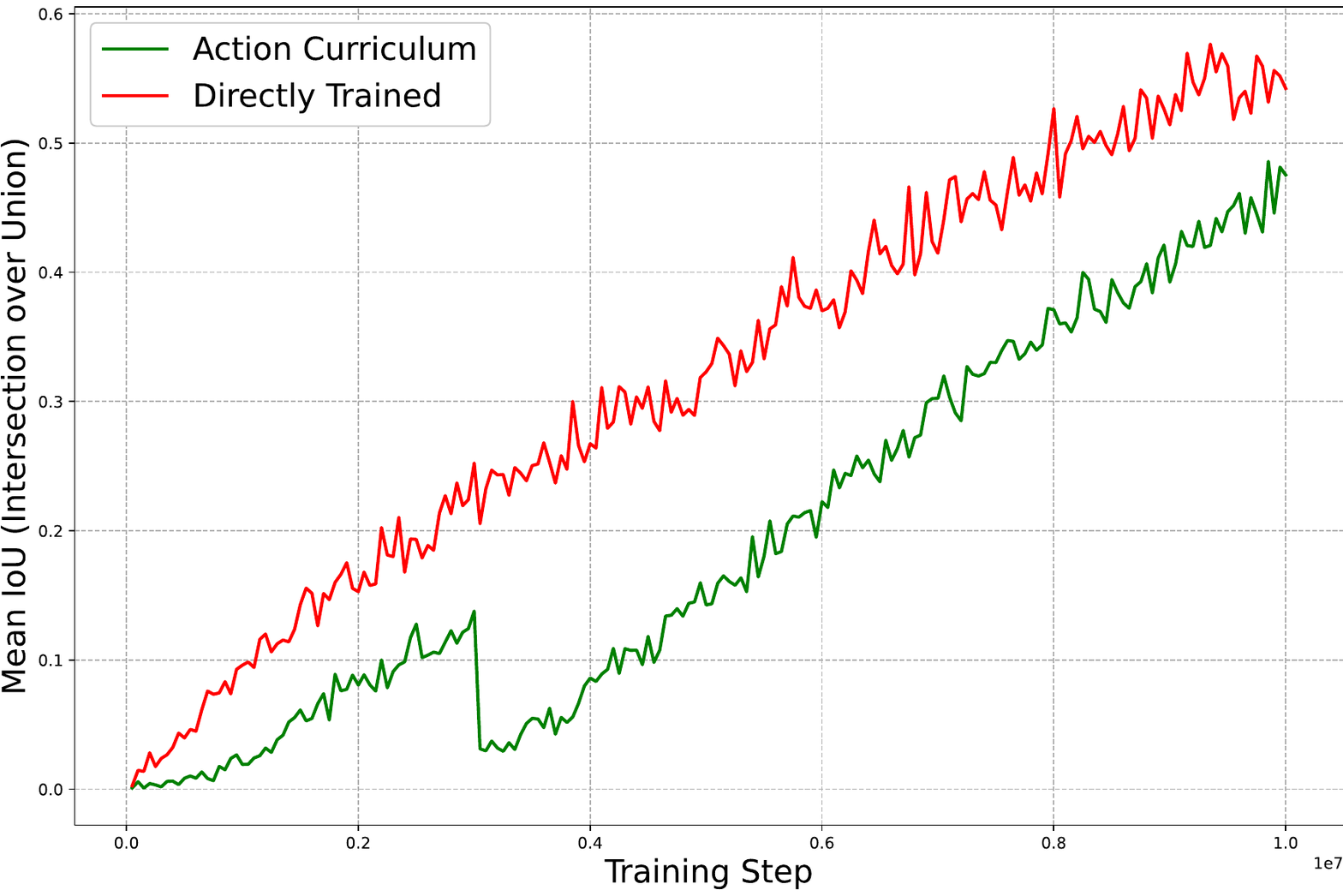}
        \caption{Action Space based Curriculum}
        \label{fig:action-curr}
    \end{subfigure}
    \hfill
    \begin{subfigure}[b]{0.27\textwidth}
        \includegraphics[width=\textwidth]{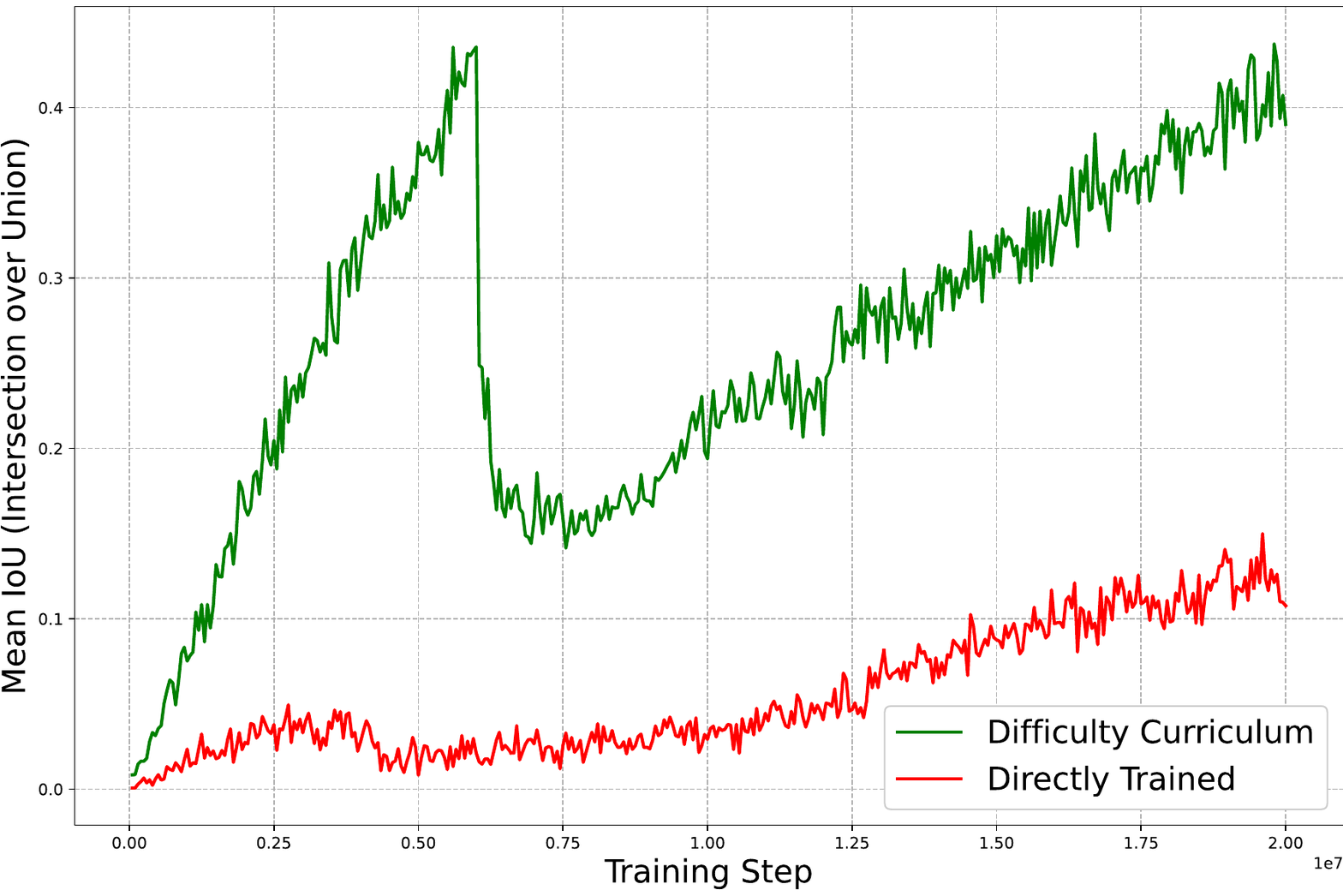}
        \caption{Difficulty Based Curriculum}
        \label{fig:diff-curr}
    \end{subfigure}
    \caption{Mean IoU of the conducted Experiments. Best viewed in colour.}
    \label{fig:curr_plots}
\end{figure}

\paragraph{Reward Functions:}
The mean IoU of all tested rewards can be seen in Fig. \ref{fig:rewards}.
Four out of the five formulated reward functions yielded satisfactory agent performance, however, using only the incremental reward (purple) did not produce any policy which was able to solve the task.
Mean IoU stayed consistently at 0.0.
The combined reward function (green)  achieved the highest mean IoU of $0.97$, followed closely by the sparse reward structure (red) with $0.96$ IoU.
Clip+ reached an IoU of $0.89$, while clip placed fourth with $0.79$ IoU.
Evident from the training graph, although reaching almost the same end performance, the combined reward achieves significantly better performance earlier in training.
Also, to achieve the 10 successful runs that are used for this comparison, we conducted 13 experiments with sparse, while only needing 11 with combined.
Explicitly, this result alone is of no statistical significance, but it does point to the combined formulation being more reliable and stable in training.

\paragraph{Curriculum Learning Experiments:}
In both curricula, the agents were able to learn even after the change in environment, as evident by the rising mean IoU after the switch. 
When increasing action space, the curriculum agent was not able to reach the performance of the normally trained agent, as seen in Fig. \ref{fig:action-curr}. 
After 10 million time steps without a curriculum (red), a mean IoU of $0.58$ was achieved. Training with the curriculum (green), the agent only achieves $0.48$.
The difficulty based curriculum yielded more promising results, which can be observed in Fig. \ref{fig:diff-curr}.
Training in the multi-detect environment from the start (red) gives a mean IoU of $0.15$ after 20 million time steps. However, using the curriculum (green), this increased to $0.43$.

Using these findings, we were able to train an agent in the multi detection environment to achieve an IoU of $0.85$ using 50 million timesteps with the same curriculum.

\section{Discussion}
The results shown in this work validate the approach and implementation of the segmentation environment.
Agents were able to utilize the information given in state and reward signals to carry out the intended task.

\paragraph{Reward Functions:}
The evaluation of the reward schemes revealed a clear favourite, that combined a step reward $r_t$, incrementally rewarding or punishing the agent based on the previous and current states, with an episodic reward $E_t$ which assesses the accomplished alignment at the end of the episode.
Solely relying on the episodic reward gave slightly worse performance, as well as noticeably worse performance earlier in training.
It was also perceived to be less reliable.
Therefore, the agent greatly benefitted from collecting the additional, step-wise reward.
Using solely the step-based reward yielded bad results, as the agents were not able to learn the task.
It seems obvious that although the step-wise reward helps when paired with the episodic reward, on its own it is not applicable.
Both clipping methods only reduced performance.
It can be concluded that the agent may achieve higher or lower returns from an episode, even when its performance aligns perfectly, depending on the initial conditions.
However, this variability does not appear to negatively impact the training process.
Furthermore, clipping seems to be detrimental to the reward signal.

\paragraph{Curriculum Learning Experiments:}
Examining the results of the curriculum experiments, in both cases agents were able to continue improving their performance after the environment was switched.
However, in the action-based experiments, the directly trained agent outperformed the curriculum agent significantly. 
While the agent does gain from the experience collected in SAT, evident by the IoU not dropping to zero after the switch, the additional time in the FAT environment seems more valuable.\\
In the difficulty-based curriculum, a slightly different picture emerges.
Here, the directly trained agent only barely learns to solve the environment, reaching a mean IoU of $0.15$.
The curriculum agent, however, greatly profits from first experiencing the single detection environment, achieving a mean IoU of $0.43$ in multi detection.
Note that directly after the switch, the performance of the curriculum agent is better than the maximum of the directly trained agent.
It indicates clearly the viability of curricula when learning a hard task that may be too demanding and too complex to learn from scratch.
Pleasingly, the agents' policy was able to adapt to the new environment without being altered too much and losing the advantage.
In conclusion, both experiments show working learning curricula.
Their applicability depend on the advantage gained in comparison to the training time lost in the environment of interest.
Moreover, when training agents to handle very difficult tasks, learning curricula are able to make these problems learnable.

We provide a novel way of detecting wire-frames in 2D projection, as well as showcase the applicability and effectiveness of RL in these types of problems.
The outlined framework and methods should be applied to more complex tasks.
\begin{multicols}{2}
{\tiny
\bibliography{citations}
}
\end{multicols}
\end{document}